%% file: icassp.tex
\documentclass{article}
\usepackage{spconf,amsmath,graphicx,hyperref}
\usepackage{booktabs}
\usepackage{multirow}
\usepackage{makecell}
\usepackage{adjustbox}
\usepackage{graphicx}
\usepackage{subcaption}

\usepackage[inline]{enumitem}

\usepackage{breqn}
\usepackage{xurl}

\title{Evaluating Machine Unlearning in ASR}
\name{Diogo Dinis$^{1,2}$, Francisco Teixeira$^1$, Bhiksha Raj$^3$, Alberto Abad$^{1,2}$, Isabel Trancoso$^{1,2}$}
\address{$^1$ INESC-ID, $^2$Instituto Superior T\'ecnico, Universidade de Lisboa, Portugal\\
$^3$LTI, Carnegie Mellon University, Pittsburgh, PA, USA}
\begin{document}
\ninept
\maketitle
\begin{abstract}
Machine unlearning (MU) offers a path to compliance with "right to be forgotten" regulations. While MU has received increasing attention for speech tasks, it remains largely unexplored for Automatic Speech Recognition (ASR). In this work, we investigate whether existing MU algorithms and evaluation tools are suitable for ASR. We apply several MU techniques to an ASR model, evaluating privacy-utility trade-offs for single-subject unlearning, then assess the best algorithm under sequential and simultaneous unlearning. Results show that gradient ascent-based algorithms achieve strong utility-privacy trade-offs, whereas more complex approaches over-unlearn samples, making them easier to identify as unlearned. This suggests standard privacy evaluations based on simple Membership Inference attacks are insufficient to reliably assess unlearning success, motivating improved evaluation methods for MU in ASR. Finally, we show that both sequential and simultaneous unlearning yield worse privacy and utility than single-subject unlearning, underscoring the need for unlearning constructions better suited to these settings.
\end{abstract}

\vspace{-0.2cm}
\section{Introduction}
\label{sec:intro}
\vspace{-0.2cm}
The widespread deployment of large-scale deep learning systems
has raised serious concerns over the privacy of training data subjects.
These concerns arise in part from the demonstrated vulnerability of such models to membership inference~\cite{shokriMembershipInferenceAttacks2017}, model inversion~\cite{hidanoModelInversionAttacks2017} and data extraction attacks~\cite{carliniExtractingTrainingData2021}, which can enable the extraction of information, or even full retrieval of training data samples (or subjects).
There is also a growing need to ensure compliance with data protection regulations worldwide, such as the European Union's General Data Protection Regulation (GDPR)~\cite{EU_2016_679}, or California's Consumer Protection Act (CCPA)~\cite{CaliforniaCCPA2018}. Among other protections, these regulations enshrine the ``right to erasure" (commonly known as the ``right to be forgotten")~\cite{EU_2016_679}, granting individuals the right to request the deletion of their personal data. 
While compliance with this protection is straightforward for stored data, this is not the case for trained model weights, which contain patterns or information related to training data that cannot be easily removed.
%

Machine unlearning (MU) techniques seeking to remove the influence of specific training samples from an already trained model 
have emerged as a potential practical alternative to full retraining~\cite{caoMakingSystemsForget2015}, not only to ensure privacy, but also as a way to minimise biases~\cite{liu-etal-2025-mitigating} or remove learned information or classes~\cite{golatkarEternalSunshineSpotless2020}. 
%
%
Following the seminal work of Cao and Yang~\cite{caoMakingSystemsForget2015}, early efforts on MU have largely focused on image classification models~\cite{bourtouleMachineUnlearning2021,golatkarEternalSunshineSpotless2020}. More recently, the growing prominence of generative models has led to an increasing number of studies on unlearning in large language models (LLMs)~\cite{dorna2026openunlearning}.
In contrast, MU for speech-based models has only recently begun to attract sustained attention. Prior research has largely focused on classification tasks, including speaker identification~\cite{chengSpeechUnlearning2025}, keyword spotting~\cite{chengSpeechUnlearning2025}, speech emotion recognition~\cite{phukanMachineUnlearningParalinguistic2025, renMachineUnlearningSpeech2025}, and spoken language understanding~\cite{koudounasAlexaCanYou2025, savelliUnSLUBENCHExtendedMachine2026,singh2026selectivecapabilityunlearningendtoend}. Beyond these, MU for text-to-speech synthesis (TTS) has also started to garner interest~\cite{kim2025do, leeErasingYourVoice2026}.

However, a significant gap remains in MU research concerning Automatic Speech Recognition (ASR).  
This gap is particularly notable given that ASR systems are among the most widely deployed speech-based models and, consequently, among the most likely targets of privacy attacks and user data deletion requests, along with TTS models.
In addition, ASR models have been shown to memorize and leak training data under certain conditions~\cite{shejwalkar24_interspeech, wangUnintendedmemorisationLarge2024}. Architectural improvements have further compounded these vulnerabilities by enabling models to be prompted, thereby expanding their attack surface. As ASR model architectures progressively move towards Speech Language Models (SLMs), which couple LLMs with pre-trained speech encoders to perform recognition, these risks are likely to increase~\cite{züfle2026helpfulcontextleaksprivacy}.
Notable efforts 
include the works of Liu~\cite{liuUnlearningLLMbasedSpeech2025} and Shamsian et al.~\cite{shamsianGoYourMeans2026}. The former evaluates Gradient Ascent~\cite{liuUnlearningLLMbasedSpeech2025} unlearning over synthetic canaries -- artificially created samples injected into the training data to act as privacy tracking devices -- providing relevant insights into memorisation in LLM-based ASR. However, the fact that canaries come from a synthetic distribution may overestimate the ability of the MU algorithm to remove the influence of data samples, as the canaries and training data will belong to different distributions. Shamsian et al.'s work, on the other hand, compares several MU methods over several tasks, including ASR, but striving to obtain an error as high as possible on the forget set, which, as we argue below, is different from our definition of~MU.

In this work, we aim to further explore the application and adaptation of MU techniques to an end-to-end ASR model. We evaluate unlearning in terms of privacy and utility, following a narrow interpretation of unlearning: the unlearned model should behave similarly to a ``gold standard" model trained from scratch without the data to be forgotten (the forget set).
This implies that it should not be possible to distinguish between a model's behaviour for an unlearned sample from a test sample drawn from the same data distribution. 

Our results show that gradient ascent-based algorithms achieve strong privacy-utility trade-offs under standard metrics. On the other hand, more complex algorithms match these trade-offs in simple MI attacks, but leave unlearned members identifiable to unlearning-informed MI attacks, indicating that naïve MI evaluations are insufficient to assess unlearning success. Moreover, the Earth mover's distance between forget-loss distributions of the unlearned and retrained (gold-standard) models remains high, suggesting unlearning does not necessarily move models in the desired direction.
Finally, sequential and simultaneous unlearning both yield worse privacy-utility trade-offs than single-subject unlearning, underscoring the need for constructions better suited to these settings.

The main contributions of this work are as follows:
\begin{itemize}
\item We conduct the first in-depth exploration of MU applied to ASR following a narrow unlearning formulation;
\item We provide a reproducible experimental setup (including a codebase) which includes five MU methods;
\item We thoroughly evaluate privacy through membership inference attacks (MIAs) in unlearning-unaware and unlearning-informed settings, and show that several MU algorithms do not hold the same privacy guarantees across the two;
\item We assess unlearning under sequential and simultaneous multi-subject scenarios, showing that both settings degrade privacy and utility relative to single-subject unlearning.
\end{itemize}

\vspace{-0.3cm}
\section{Machine Unlearning in ASR}
\vspace{-0.1cm}
\subsection{Problem statement}
\label{sec:background}

 
Let $f_\theta:\mathcal{X}\rightarrow\mathcal{Y}$ be an ASR model, parametrised by $\theta = \mathcal{A}(f, \mathcal{D}_{train})$, where $\mathcal{A}$ is a randomised supervised learning algorithm, and $\mathcal{D}_{train}=\left\{\left(x_i,y_i,s_i\right)\right\}_{i=1}^{N}$ a training set with $N$ triples of speech recordings $x_i\in\mathcal{X}$, ground-truth transcriptions $y_i\in\mathcal{Y}$, and speaker identities $s_i\in\mathcal{S}$, drawn from a data distribution $\mathcal{D}$. 
Given a deletion request from a speaker $s_f\in\mathcal{S}$, let $\mathcal{D}_f=\{(x_i,y_i,s_i)\in\mathcal{D}_{train}: s_i=s_f\}$ be the subset of speech data to be \textit{forgotten}, and $\mathcal{D}_r\subseteq \mathcal{D}_{train}\setminus \mathcal{D}_f$ the set of data to be retained, commonly called the ``forget set'' and ``retain set'', respectively. 
Informally, the goal of a MU algorithm $\mathcal{U}$ is to produce $\theta^{\mathcal{U}}$ as close as possible to $\theta^{\mathcal{G}} = \mathcal{A}(f, \mathcal{D}_{train} \setminus \mathcal{D}_f)$, with $\theta^{\mathcal{G}}$ being the ``gold standard'' model, re-trained without the forget set\footnote{Concretely, the randomised nature of $\mathcal{A}$ induces a distribution of possible models. The ``gold standard'' is therefore a distribution and the goal of unlearning is to obtain a model that is more likely to be drawn from the distribution induced by $\mathcal{A}(f, \mathcal{D}_{train} \setminus \mathcal{D}_f)$ than by $\mathcal{A}(f, \mathcal{D}_{train})$~\cite{ triantafillouYourAlgorithmUnlearning2026}.}.
An unlearned ASR model that is close to a model re-trained from scratch should not have an uncharacteristically high Word Error Rate (WER) and loss for the forgotten subject, but instead WER and loss values distributionally close to unseen data from the same distribution.

\vspace{-0.2cm}
\subsection{Machine Unlearning paradigms}\label{back:methods}
\vspace{-0.1cm}
There are two main types of unlearning techniques: exact and approximate.
Exact Unlearning is a branch of MU where algorithms are designed to provide formal guarantees that $\theta^{\mathcal{U}}$ was not trained on $\mathcal{D}_f$. 
These techniques introduce changes in the model's architecture and training process itself, such that, to unlearn a sample, it is only necessary to re-train a small part of the model at a much lower cost than full re-training. Since the resulting model will not have been trained on $\mathcal{D}_f$, it can be said that it has exactly forgotten it. For instance, the ``Sharded, Isolated, Sliced, and Aggregated'' (SISA)~\cite{bourtouleMachineUnlearning2021} algorithm splits the model into multiple replicas, each trained on a disjoint subset of $\mathcal{D}_{train}$, ensuring that the influence of any given set is confined to a single replica. This way, the retraining process for $\mathcal{D}_f$ is reduced to the smaller replicas trained on its constituents, and maintains the guarantee of the absence of $\mathcal{D}_f$ in $\theta^{\mathcal{U}}$.


Approximate Unlearning methods, on the other hand, have the goal of reducing the influence of $\mathcal{D}_f$ on $\theta$,
without retraining $f_\theta$, and without requiring specialised pre-training mechanisms. Instead, Approximate Unlearning algorithms correspond to some form of post-hoc adaptation of pre-trained models, often in the form of gradient ascent on data from the forget set in combination with finetuning on the retain set, to ensure model utility is kept. In this case, however, existing algorithms do not provide exact guarantees of unlearning, and unlearning success in terms of privacy must be evaluated empirically.
Nevertheless, while exact methods provide formal unlearning guarantees, their reliance on specific training algorithms makes them unsuitable for existing deployed models.

\vspace{-0.2cm}
\subsection{Evaluation}\label{evaluation_intro}
\vspace{-0.1cm}
Machine unlearning algorithms need to be evaluated at two levels: utility, to ensure that the unlearning process kept the model's performance; and unlearning success. Evaluating utility is straightforward, as the model's performance metrics are usually well established beforehand. 
In contrast, evaluating unlearning success depends heavily on the unlearning objective.
For privacy, unlearning success is most commonly measured in terms of a membership inference (MI) attacker's success in correctly identifying unlearned samples as part of the training set~\cite{shokriMembershipInferenceAttacks2017}. Membership inference attacks act as a proxy for how much a model has memorised or overfitted to a sample. Other possible measures of unlearning success include data extraction attacks that attempt to retrieve information about the forget set~\cite{carliniExtractingTrainingData2021}, and statistical indistinguishability tests between unlearned and retrained models~\cite{zhang2026unlearning}. Nevertheless, data extraction attacks are not well developed for all applications, whereas statistical indistinguishability tests can become computationally prohibitive for large models~\cite{triantafillou2024we}.

\vspace{-0.1cm}
\subsection{Proposed Machine Unlearning methods for ASR}
\vspace{-0.15cm}
In this work, we adapt and evaluate the following approximate MU methods for ASR, given their wider applicability:

\vspace{0.1cm}
\noindent\textbf{Baseline -- Finetuning:}
This method performs gradient descent on $\mathcal{D}_r$ to reduce the influence of $\mathcal{D}_f$ on the model, deliberately overfitting $\mathcal{D}_r$ to induce catastrophic forgetting in $\mathcal{D}_f$.

\vspace{0.05cm}
\noindent\textbf{Baseline -- CF-$k$:}
Unlike simple finetuning, this method, ``Catastrophically forgetting the last $k$ layers''~\cite{goelAdversarialEvaluationsInexact2023}, focuses only on finetuning the last $k$ layers, freezing the preceding layers.

\vspace{0.05cm}
\noindent\textbf{NegGrad and NegGrad+:}
NegGrad~\cite{golatkarEternalSunshineSpotless2020}, or Gradient Ascent, is the most common unlearning method. The model is finetuned with $\mathcal{D}_f$, using the reverse of the gradient direction, effectively moving the model's weights in the direction of increasing loss for these samples. A popular extension of this method, NegGrad+~\cite{kurmanjiUnboundedMachineUnlearning2023}, mitigates catastrophic forgetting by additionally finetuning on $\mathcal{D}_r$. 

\vspace{0.05cm}
\noindent\textbf{SCRUB:}
``SCalable Remembering and Unlearning unBound''~\cite{kurmanjiUnboundedMachineUnlearning2023}, or SCRUB, uses a teacher-student setup, with the original model 
as a teacher. This method uses three different losses: a distillation loss, instantiated as the Kullback-Leibler (KL) divergence, aiming to maximise the similarity between the student and the teacher on the retain set, $\mathcal{D}_r$; 
a task loss, which is applied only to $\mathcal{D}_r$; and the negative KL divergence, which aims to minimise the similarity between the teacher and student on $\mathcal{D}_f$.

\vspace{0.05cm}
\noindent\textbf{ASU:}
Attention Smoothing Unlearning~\cite{zadeAttentionSmoothingAll2025} also uses a teacher-student setup, using the original model as a teacher, increasing the Softmax temperature in the teacher's attention layers, smoothing the attention distributions, producing less confident outputs on $\mathcal{D}_f$. 
The student is trained to minimise the KL divergence between its and the smoothed teacher's outputs on $\mathcal{D}_f$. 
Our adaptation of this method for ASR targets the 
attention layers of the ASR model's decoder.




\vspace{-0.2cm}
\section{Experimental Setup}\label{exp_setup}

\vspace{-0.1cm}
\subsection{Model selection and implementation}
\vspace{-0.1cm}
For the experiments in this work, we selected a state-of-the-art, open-source, pre-trained E-Branchformer~\cite{kimEBranchformerBranchformerEnhanced2023}\footnote{\url{https://huggingface.co/asapp/e\_branchformer\_librispeech}} to ensure a fully transparent pipeline and, most importantly, strict traceability of the datasets and data partitions used for training. 
This model was trained with the LibriSpeech ASR recipe from ESPnet~\cite{watanabe18_interspeech}, which uses the full 960 hours of training data from LibriSpeech~\cite{panayotovLibrispeechASRCorpus2015}. 
\noindent\textbf{Retrained Model:}
In order to obtain a gold standard, we retrained the full model from scratch. Although such retraining would preferably be done once for each of the subjects to be unlearned, to minimise computational costs, we opted to retrain a single model using the original model's training set (LibriSpeech's ``train-960''), and excluding the 10 forget subjects at the same time.

All experiments were performed on a single computation node with an Intel(R) Xeon(R) Gold 6348 CPU and 3 NVIDIA RTX A6000 GPUs. We make our codebase openly available on GitHub\footnote{\url{https://github.com/dmbdpt/asr-unlearning-evaluation}}. 

\vspace{-0.3cm}
\subsection{Data}\label{sec:data}
\vspace{-0.2cm}
Three partitions of LibriSpeech were used in our experiments: ``train-clean-100'', ``test-clean'', and ``test-other''. Each was broken into different subsets for unlearning and evaluation. 
\input{tables/data_part_LS}
%
We designed our experiments as speaker-level unlearning tasks. For a target speaker $s_f$, the forget set, $\mathcal{D}_f$ contains all of that speaker's utterances in ``train-clean-100'', while the retain set, $\mathcal{D}_r$ includes all utterances from the remaining subjects in that set. While MU implementations often set $\mathcal{D}_r$ as $\mathcal{D}_{train}\setminus \mathcal{D}_f$, we used
``train-clean-100'' as a representative subset of the model's full training set (LibriSpeech's ``train-960'').
In total, 10 pairs of \textit{forget/retain} sets were generated, corresponding to 10 speakers to be forgotten. We reserved an additional speaker, disjoint from the previous ten, for hyperparameter search.
The aforementioned data partitions are detailed in the first half of Table~\ref{tab:data_part_ls}.

\vspace{-0.2cm}
\subsection{Hyperparameter selection}
\vspace{-0.1cm}
To better compare the chosen unlearning methods, we standardised the unlearning process to 10 epochs with a fixed batch size of 8. 
%
We determined the remaining hyperparameters through a Bayesian search using the Optuna~\cite{akibaOptunaNextgenerationHyperparameter2019} library's default algorithm, a Tree-structured Parzen Estimator sampler.
Considering $\mathcal{L}_M(S)$ as the distribution of per-utterance losses of model $M$ over set $S$, our objective was defined as the minimisation of the Earth Mover's Distance (EMD), 
between the loss distributions of the forget set and the pre-unlearning test set, i.e., $\text{EMD}(\mathcal{L}_{\theta^U}({\mathcal{D}_f}), \mathcal{L}_{\theta}({\mathcal{D}_{test}}))$.
\vspace{-0.3cm}

\subsection{Utility evaluation}
\vspace{-0.1cm}
To evaluate utility, we compute the
mean WER and model loss values, corresponding to the Connectionist Temporal Classification (CTC) loss from the model's encoder output and the cross-entropy (CE) loss computed over the decoder's output, for the forget, retain, and test sets, for each speaker to forget. We include the loss, as a direct measure of the alignment between the model's behaviour on the forget set, and its performance on the retain and test sets.
We use the model's default decoding configuration, except for the beam size, which is set to 5. 

\vspace{-0.2cm}
\subsection{Privacy evaluation}
\vspace{-0.1cm}
To evaluate unlearning success in terms of privacy, we created data partitions and implemented two types of MI attacks: \textit{simple} and \textit{informed}. Both are evaluated in terms of Area Under the Curve (AUC) and Equal Error Rate (EER).
We additionally include the $\text{EMD}(\mathcal{L}_{\theta^U}({\mathcal{D}_f}), \mathcal{L}_{\theta^U}({\mathcal{D}_r}))$,  $\text{EMD}(\mathcal{L}_{\theta^U}({\mathcal{D}_f}), \mathcal{L}_{\theta^U}({\mathcal{D}_{test}}))$ and $\text{EMD}(\mathcal{L}_{\theta^U}({\mathcal{D}_f}), \mathcal{L}_{\theta^{\mathcal{G}}}({D_{f}}))$, representing how close the loss distribution over the forget set is to the corresponding model's retain and test sets, as well as to the forget set's loss distribution on the re-trained model. These metrics are used to gain a more in-depth understanding of the results obtained for the MI attackers.

\noindent\textbf{Attack partitions:} MI attacks require observed and non-observed data (members and non-members). The attacker is trained on utterances from the $\mathcal{D}_r$ as positives (\textit{members\_train}) and from $\mathcal{D}_{test}$ as negatives (\textit{non-members\_train}), and subsequently evaluated on $\mathcal{D}_f$ as positives (\textit{members\_test}) against held-out samples from $\mathcal{D}_{test}$ as negatives (\textit{non-members\_test}). The evaluation asks whether the attack recognises the forgotten sample as a member, which, for a perfectly unlearned model, should yield an AUC of 50\%.
After observing in initial experiments that mean utterance duration varies substantially between the original train and test sets, making duration a confounding factor for MI attacks, MI partitions were matched by duration.
Further details of the partitions are provided in the second half of Table~\ref{tab:data_part_ls}.

\noindent\textbf{Simple Attacker:} employs a Random Forest (RF) classifier trained with the target models' encoder's CTC loss and decoder's cross-entropy loss~\cite{teixeiraExploringFeaturesMembership2026}.
The RF is trained on utterances from \textit{members\_train} and \textit{non-members\_train}, and evaluated on \textit{members\_eval} and \textit{non-members\_eval}. 
However, this attack has no reference for how the model behaves with regard to an unlearned sample. As such, if the unlearned samples' losses diverge from the retain samples' losses (e.g., by having a very high loss), this attacker might fail to recognise them as belonging to the original training set.
\input{tables/results_test_clean_other_neg_MI}

\noindent\textbf{Informed Attacker:} follows the \textit{simple attacker}, but attempts to address its na\"ive construction. Specifically, the informed MI attacker leverages all unlearned models to train the RF classifier in a leave-one-out fashion -- i.e., for each forget subject, we leverage the remaining 9 other subjects' unlearned model losses as an ``ìnformed'' \textit{members\_train}, along with the losses for the same models over \textit{non-members\_train}, to train the RF; and then use the target forget and test subject samples' losses computed on the unlearned model for evaluation. 
This way, the classifier is able to learn the behaviour of a model with regard to forget samples after unlearning.

\vspace{-0.2cm}
\section{Results}
\vspace{-0.1cm}

The results for the application of the MU algorithms of Section \ref{back:methods} to ASR can be found in Table~\ref{tab:results}. 

\noindent\textbf{Utility:} We observe that NegGrad, NegGrad+, SCRUB and AttSmooth have the closest WER values to $\theta^{\mathcal{G}}$, for all sets. On the other hand, in all cases, the two  baseline methods (finetune and CF-k) cause noticeable overfitting to the retain set and markedly degrade performance in both test partitions. 
In terms of mean loss distributions, the algorithms behave in the same way as for the WER for all sets except for the $\mathcal{D}_f$, wherein both versions of NegGrad and SCRUB highly degrade the loss. 


\noindent\textbf{Privacy:} Using the simple MIA attack, results behave similarly to those observed for utility. Finetune and CF-k have the worst performances in terms of privacy, whereas the remaining methods present strong privacy improvements. The best performances are achieved by NegGrad+ and AttSmooth. 
Note that, while NegGrad, NegGrad+ and SCRUB have higher losses, this does not translate into worse privacy results. 
This comes from the fact that the simple MI attack is trained to distinguish between train and test losses, which means that, while these MU algorithms may increase $\mathcal{D}_f$'s loss by a large amount, $\mathcal{D}_f$'s loss distribution may still be closer to the test set than to $\mathcal{D}_r$. 
This is validated by the EMD
columns reported in the table, where, in all cases that the EMD between $\mathcal{L}_{\theta^U}(D_f)$ and $\mathcal{L}_{\theta^U}(D_{test})$ is smaller than the EMD between $\mathcal{L}_{\theta^U}(D_f)$ and $\mathcal{L}_{\theta^U}(D_r)$,
the MI classifier has very poor performance.

\noindent\textbf{Informed attacker:} Regarding the informed attacker's results, we observe a very large privacy degradation for both SCRUB and AttSmooth, with the informed MI attack achieving an AUC of 87.57\% for AttSmooth, compared to the 52.09\% obtained by the simple MI attacker.
To understand this large discrepancy, we analysed the classifier's decision boundary for AttSmooth. We found that while the simple attacker is able to rely on the combination of both losses to identify members and non-members, the informed attacker, trained with unlearned samples, relies much more strongly on higher CE losses to predict members (unlearned samples). This is likely due to the fact that smoothing is only applied in decoder layers, making this loss much higher for unlearned samples.

If we consider only MI and WER, NegGrad+ has the strongest  privacy and utility trade-off, closely followed by NegGrad. Nevertheless, NegGrad is the more computationally efficient alternative, as it only requires gradient ascent on $\mathcal{D}_f$, contrary to NegGrad+, that also finetunes on $\mathcal{D}_r$.
However, we also note that the lowest EMD($\mathcal{L}_{\theta^U}(\mathcal{D}_f)$, $\mathcal{L}_{\theta^{\mathcal{G}}}(\mathcal{D}_f))$ are obtained by finetuning and CF-k. This means that, while some of the tested unlearning methods prevent successful MIAs, they are not yielding models that behave exactly like the re-trained model, which is the true goal of unlearning. 

\noindent\textbf{Sequential and simultaneous unlearning}: 
In a real-world scenario, deletion requests might accumulate or come in batches. We therefore evaluated the two gradient ascent-based methods by unlearning subjects sequentially and simultaneously, using the hyperparameters selected for single-subject unlearning. The results are presented at the end of Table~\ref{tab:results}. 
Under the sequential approach, NegGrad severely degrades the model, with the retain WER rising to 2.93\% against 0.41\% on the retrained model. Although its MIA AUC of 59.39\% is the highest among the multi-subject runs, the low EMD between $\mathcal{L}_{\theta^U}(\mathcal{D}_f)$ and $\mathcal{L}_{\theta^U}(\mathcal{D}_r)$ (0.86) provides no evidence of forget-set memorisation, instead indicating convergence of the forget and retain distributions.
NegGrad+ improves on this metric through the retain finetuning, and its forget-set metrics are the closest to $\theta^{\mathcal{G}}$ of all four multi-subject runs. Even so, it still degrades test performance, with a Test-Other WER of 6.81\% against 4.25\% and 5.02\% on the single-subject case and retrained model, respectively, and a rise in MIA AUC from 50.18\% to 55.58\% when compared to the single-subject case.
Under simultaneous unlearning, the two methods fail in opposite directions. NegGrad fails to unlearn, with a forget WER of 0.55\% and loss of 3.16, close to the original model (0.40\% and 1.22), as does the MIA AUC (65.28\%), suggesting that sharing the gradient ascent over ten speakers dilutes the per-speaker signal. NegGrad+, by contrast, over-unlearns, reaching a forget loss of 14.91 and the largest distance to $\theta^{\mathcal{G}}$ in the table, with an EMD of 10.18.
Overall, the single-subject hyperparameters do not transfer reliably to multi-subject unlearning.



\noindent\textbf{Limitations:} Even though the current work focuses on a single ASR model trained on a single dataset, we consider that this is sufficient to validate the conclusions that can be taken from this paper. Nevertheless, we believe that future work should focus on extending the experiments conducted in this paper to other model architectures and datasets. Membership inference attacks were also limited to simple loss classification, which had a weak original performance ($\sim$67\%). More complex constructions exist in the literature~\cite{tao2026information,jebreel2026revisitingliramembershipinference} and should be explored in the context of MU for ASR.

\vspace{-0.1cm}
\section{Conclusions}
\vspace{-0.1cm}
In this paper, we focused on studying and evaluating MU algorithms for ASR. Overall, our simple MI-based evaluation results identified four algorithms (NegGrad, NegGrad+, SCRUB and AttSmooth) with strong trade-offs between privacy and utility.
However, under unlearning-informed MI attacks, SCRUB and AttSmooth were found to produce loss patterns that allowed the identification of forgotten samples. Results for sequential and simultaneous unlearning, which mimic real-world conditions, were also found to degrade both privacy and utility. 
In addition, our results showed an inversion between EMD and MIA: the methods most resistant to MIA were farthest from the gold standard in terms of forget-set loss distribution, while those closest to it were most vulnerable. Resistance to MIAs therefore does not immediately imply absence of information about the forget data in the unlearned models.
We therefore consider that MU-informed MIA constructions are necessary for a thorough evaluation of privacy, while distributional measures should be included in evaluation setups to further validate similarity to gold standard models.
This study highlights MU as a compelling and critical open problem. By benchmarking MU algorithms and exposing flaws in standard evaluation metrics, this work tried to provide a stepping stone for future research in MU for ASR and other speech tasks.

\section{Acknowledgments}
This work was supported by national funds through Fundação para a Ciência e a Tecnologia, I.P. (FCT) under projects
UID/50021/2025, UID/PRR/50021/2025 and CMU-Portugal project \url{https://doi.org/10.54499/2024.14611.CMU} (LeaF).

\bibliographystyle{IEEEbib}
\bibliography{mybib}

\end{document}

%% file: tables/data_part_LS.tex
\begin{table}[t]
\centering
\caption{Data used for model training, unlearning and utility evaluation and data partitions for MI evaluation. \#Spk. and \#Utt. correspond to the average over the partitions for all 10 forget subjects.}
\vspace{-0.2cm}
\label{tab:data_part_ls}
\resizebox{\columnwidth}{!}{
\begin{tabular}{l c c c c}
\toprule
\textbf{Partition} & \textbf{\#Spk.} & \textbf{\#Utt.} & \textbf{Avg. Dur. (s)} & \textbf{Source (LibriSpeech)} \\\midrule
\multicolumn{5}{l}{\textbf{Unlearning \& Utility Eval.}} \\
 \textit{forget}     & 1   & 105    & 11.9 & \textit{train-clean-100} \\
 \textit{retain}     & 250 & 28,434 & 12.7 & \textit{train-clean-100} \\
 \textit{test-clean} & 40  & 2,620  & 7.4  & \textit{test-clean} \\
 \textit{test-other} & 33  & 2,939  & 6.5  & \textit{test-other} \\
 \textit{test}    & 73  & 5,559  & 7.0  & \textit{test-clean} \& \textit{test-other} \\ 
\midrule
\multicolumn{5}{l}{\textbf{Membership Inference Eval.}} \\
\textit{members\_train}    & 219.8 & 736.6 & 11.9 & \textit{train-clean-100} \\
\textit{non-members\_train} & 29.9  & 82.9  & 11.8 & \textit{test-clean} \& \textit{test-other} \\
\textit{members\_eval}     & 1     & 105   & 11.9 & \textit{train-clean-100} \\
\textit{non-members\_eval}  & 29.4  & 82.0  & 11.9 & \textit{test-clean} \& \textit{test-other} \\
\bottomrule
\end{tabular}}
\vspace{-0.3cm}
\end{table}

%% file: tables/results_test_clean_other_neg_MI.tex
\begin{table*}[ht]
\centering
\caption{Main utility and MIA results. WER and Loss should be close to the re-trained model. AUC and EER should be close to 50\%. The EMD between $\mathcal{L}_{\theta^U}({\mathcal{D}_f})$, $\mathcal{L}_{\theta^U}({\mathcal{D}_{test}})$ and $\mathcal{L}_{\theta^{\mathcal{G}}}({D_{f}})$ should be low, while EMD between $\mathcal{L}_{\theta^U}({\mathcal{D}_f})$ and $\mathcal{L}_{\theta^U}({\mathcal{D}_r})$ should be high.} 
\vspace{-0.1cm}
\label{tab:results}
\begin{adjustbox}{max width=\textwidth}
\begin{tabular}{l | cc cc cc cc || c c c || cc | cc }
\toprule
& \multicolumn{2}{c}{\textbf{Forget ($\mathcal{D}_f$)}}
& \multicolumn{2}{c}{\textbf{Retain ($\mathcal{D}_r$)}}
& \multicolumn{2}{c}{\textbf{Test-Clean}}
& \multicolumn{2}{c||}{\textbf{Test-Other}}
& \multicolumn{3}{c||}{\textbf{EMD (vs $\mathcal{L}_{\theta^U}({\mathcal{D}_f})$)}}
& \multicolumn{2}{c|}{\textbf{MIA}} 
& \multicolumn{2}{c}{\textbf{IMIA}} \\
\cmidrule(lr){2-3}
\cmidrule(lr){4-5}
\cmidrule(lr){6-7}
\cmidrule(lr){8-9}
\cmidrule(lr){10-12}
\cmidrule(lr){13-14}
\cmidrule(lr){15-16}
\textbf{Method}
& WER (\%) & Loss
& WER (\%) & Loss
& WER (\%) & Loss
& WER (\%) & Loss
& \multicolumn{1}{c}{\multirow{1}{*}{\textbf{($\mathcal{L}_{\theta^U}({\mathcal{D}_r})$)}}}
& \multicolumn{1}{c}{\multirow{1}{*}{\textbf{($\mathcal{L}_{\theta^U}({\mathcal{D}_{test}})$)}}}
& \multicolumn{1}{c||}{\multirow{1}{*}{\textbf{($\mathcal{L}_{\theta^{\mathcal{G}}}({D_{f}})$)}}}
& AUC (\%) & EER (\%) 
& AUC (\%) & EER (\%) \\
\midrule
Original        & 0.40            & 1.22            & 0.46          & 1.19          & 1.89          & 3.04           & 4.00          & 5.09          & 0.31  & 3.17 & 3.26 & 67.48         & 36.45           & --              & --              \\
Re-trained      & 2.01             & 4.51             & 0.41        & 1.58          & 2.18          & 3.22           & 5.02          & 5.33          & 3.03  & 2.56 &  --  & 51.97         & 49.70           & --              & --              \\ \midrule
Finetune        & 1.84            & \textbf{4.46}   & 0.16          & 0.76          & 3.54          & 6.17           & 9.80          & 13.24         & 4.28 & 5.06 & 1.73 & 68.01          & 36.16           & 58.45           & 43.46           \\
CF-k            & 1.45            & 3.78            & 0.14          & 0.77          & 3.01          & 5.49           & 6.11          & 9.54          & 3.49 & 3.66 & 1.10 & 63.87          & 39.86           & 56.81           & 46.08           \\
NegGrad         & \textbf{2.01}   & 6.62            & 0.51          & 2.89          & 1.99          & 4.24           & 4.21          & 6.30          & 4.28 & 3.58 & 3.00 & 52.13          & 47.67           & 53.95           & \textbf{47.09}  \\
NegGrad+        & 2.36            & 10.36           & 0.55          & \textbf{1.61} & \textbf{2.03} & \textbf{3.41}  & 4.25          & \textbf{5.57} & 9.79 & 7.58 & 8.91 & \textbf{50.18} & 49.35           & \textbf{52.69}  & 46.60           \\
SCRUB           & 1.94            & 6.48            & \textbf{0.42} & 0.97          & 1.94          & 2.97           & 4.08          & 5.01          & 6.13 & 4.19 & 4.23 & 46.00          & 51.78           & 68.98           & 35.87           \\
AttSmooth       & 2.09            & 3.83            & 0.13          & 0.58          & \textbf{2.03} & \textbf{3.41}  & \textbf{4.32} & 5.72          & 3.19 & 3.14 & 2.41   & 52.09        & \textbf{50.29}  & 87.57           & 20.00           \\
\midrule

NegGrad Seq.  & 3.82 & 15.11& 2.93 & 12.60& 4.77 & 9.80 & 8.84 & 13.16 & 0.86 & 3.39 & 2.78 & 59.39 & 42.02 & 62.30 & 41.07 \\
NegGrad+ Seq. & 2.01 & 4.70 & 0.75 & 2.07 & 3.39 & 3.84 & 6.81 & 6.28 & 2.69 & 2.76 & 3.69 & 55.58 & 45.20 & 56.32 & 44.77 \\
\midrule
NegGrad Sim.  & 0.55 & 3.16  & 0.45 & 3.03 & 2.66 & 4.01 & 5.80 & 5.95 & 0.61 & 3.43 & 2.80 & 65.28 & 37.40 & 67.45 & 38.51 \\
NegGrad+ Sim. & 3.32 & 14.91 & 1.22 & 4.63 & 4.00 & 5.79 & 7.53 & 8.45 & 10.13 & 8.69 & 10.18 & 51.57 & 48.35 & 58.47 & 43.80 \\

\bottomrule
\end{tabular}
\end{adjustbox}
\vspace{-0.2cm}
\end{table*}